\pdfoutput=1

\documentclass[11pt]{article}

\usepackage[preprint]{acl}

\usepackage{times}
\usepackage{latexsym}

\usepackage[T1]{fontenc}

\usepackage[utf8]{inputenc}

\usepackage{microtype}

\usepackage{inconsolata}

\usepackage{graphicx}

\usepackage{booktabs}
\usepackage{multirow}
\usepackage{colortbl}
\usepackage{xcolor}
\usepackage{hyperref}
\usepackage{caption}
\usepackage{subcaption}
\usepackage{amsmath}
\usepackage[textsize=tiny]{todonotes}
\usepackage{hyperref}       
\usepackage{url}            
\usepackage{booktabs}       
\usepackage{amsfonts}       
\usepackage{nicefrac}       
\usepackage{microtype}      
\usepackage[ruled]{algorithm2e}
\usepackage{graphicx}
\usepackage{subcaption}
\usepackage{makecell}
\usepackage{threeparttable}

\definecolor{darkgreen}{HTML}{92D050}
\definecolor{lightgreen}{HTML}{E0F1CB}
\title{AXCEL: Automated eXplainable Consistency Evaluation using LLMs}

\author{{\bf P Aditya Sreekar}\thanks{These authors contributed equally to this work.} \\ Amazon \\ \texttt{sreekarp@amazon.com} \And {\bf Sahil Verma}\footnotemark[1] \\ Amazon \\ \texttt{vrsahil@amazon.com} \And {\bf Suransh Chopra} \\ Amazon \\ \texttt{suransh@amazon.com} \AND {\bf Sarik Ghazarian} \\ Amazon \\ \texttt{sghazari@amazon.com} \And {\bf Abhishek Persad} \\ Amazon \\ \texttt{persadap@amazon.com} \And {\bf Narayanan Sadagopan} \\ Amazon \\ \texttt{sdgpn@amazon.com}
}

\begin{document}
\newcommand{\sahil}[1]{\textcolor{red}{#1}}
\newcommand{\sreekar}[1]{\textcolor{blue}{#1}}
\maketitle

\begin{abstract}

Large Language Models (LLMs) are widely used in both industry and academia for various tasks, yet evaluating the consistency of generated text responses continues to be a challenge. Traditional metrics like ROUGE and BLEU show a weak correlation with human judgment. More sophisticated metrics using Natural Language Inference (NLI) have shown improved correlations but are complex to implement, require domain-specific training due to poor cross-domain generalization, and lack explainability. More recently, prompt-based metrics using LLMs as evaluators have emerged; while they are easier to implement, they still lack explainability and depend on task-specific prompts, which limits their generalizability. This work introduces \textbf{A}utomated e\textbf{X}plainable \textbf{C}onsistency \textbf{E}valuation using \textbf{L}LMs (AXCEL), a prompt-based consistency metric which offers explanations for the consistency scores by providing detailed reasoning and pinpointing inconsistent text spans. AXCEL is also a generalizable metric which can be adopted to multiple tasks without changing the prompt. AXCEL outperforms both non-prompt and prompt-based state-of-the-art (SOTA) metrics in detecting inconsistencies across summarization by 8.7\%, free text generation by 6.2\%, and data-to-text conversion tasks by 29.4\%. We also evaluate the influence of underlying LLMs on prompt based metric performance and recalibrate the SOTA prompt-based metrics with the latest LLMs for fair comparison. Further, we show that AXCEL demonstrates strong performance using open source LLMs. 

\end{abstract}
\section{Introduction}
\label{sec:introduction}

Recent advances in natural language processing (NLP) have led to the development of Large Language Models (LLM), which can generate text virtually indistinguishable from that created by humans. These models are adept at interpreting and executing natural language instructions \citep{ouyang2022training, rafailov2023direct}, allowing them to undertake tasks such as summarization and classification directly, without additional training \citep{brown2020language}. This has catalysed their incorporation into a broad spectrum of mainstream applications, including summarization systems, interactive chatbots, and virtual assistants. However, studies \citep{bang2023multitask,raunak-etal-2021-curious} have shown that LLM outputs can contain hallucinations, meaning the output is not consistent or irrelevant to the given context. 
Identifying these consistency issues is challenging because the errors typically align with the task's overarching structure and theme, making them subtle and difficult to detect. Evaluating the generated text for such issues is crucial to ensure trust and reliability, enabling successful adoption of LLMs across various applications.

A widely accepted gold-standard approach for evaluating generated text is human evaluation. However, it is labor-intensive, time-consuming, and requires subject-matter expertise, making it expensive for large-scale applications. This underscores the need for automated metrics to efficiently evaluate consistency, \textit{defined as the degree to which the generated text aligns with its source}. In the context of this paper, consistency and hallucination are inversely related; lower consistency scores correspond to higher degrees of hallucination. Thus, inconsistency and hallucination are used interchangeably. Several metrics have been proposed in the literature for measuring consistency, ranging from simple textual similarity-based metrics to Natural Language Inference (NLI) based methods. Textual similarity-based metrics, such as ROUGE \citep{lin-2004-rouge}, BLEU \citep{papineni-etal-2002-bleu}, and BERTScore \citep{zhang2019bertscore}, require a reference text written by humans and exhibit weak correlation with human evaluation \citep{liu-etal-2023-g, zha2023alignscore}. On the other hand, NLI-based methods, like AlignScore \citep{zha2023alignscore} and Summac \citep{laban2021summac}, are reference-free and demonstrate better correlation with human judgment; however they struggle with generalization and require bespoke models trained on specific data to enhance generalization. Recent interest in LLMs has led to the application of LLMs as evaluators \citep{liu-etal-2023-g, fu2023gptscore, gao2023humanlike, chen2023evaluating}. These LLM based approaches design prompts to assess the factual consistency of generated texts with respect to the source text. These prompt-based metrics show strong correlation with human judgments and do not require custom training, relying instead on the emergent abilities of LLMs. However, their effectiveness is limited by the specificity of their prompts, which are often tailored to particular tasks, such as GEval\cite{liu-etal-2023-g} using different prompts for different datasets. Additionally, both NLI-based and prompt-based metrics are black-box evaluators, providing scores without explanations. 
Explainability is crucial not only for building trust in the metrics, but also for providing actionable feedback to improve upstream text generation systems.

To address these challenges, we introduce \textbf{A}utomated e\textbf{X}plainable \textbf{C}onsistency \textbf{E}valuation using \textbf{L}LMs (AXCEL), a reference-free, prompt-based metric that exhibits a high level of agreement with human judgment and provides explainability. AXCEL utilizes Chain of Thought (CoT) and few shot prompting techniques. Beyond scoring, AXCEL also provides detailed reasoning behind the score, facilitating more trust and easier identification of hallucinations in consistency scoring as compared to black box approaches. AXCEL also generalises across multiple tasks and LLMs of varied sizes without requiring any modifications to the prompt. This capability stems from the in-context learning via the few shot exemplars which allow it to get an understanding of the tasks.  
In experiments, AXCEL is benchmarked on three tasks: 1) summarization, a typical NLP task with textual inputs and outputs; 2) free text generation, a setting without any external knowledge source to evaluate consistency of generated text; 3) data-to-text conversion, conversion of JSON to a textual overview. These tasks were selected due to the diversity they represent, aiding in evaluating the generalizability of AXCEL. Further, prompt-based metrics heavily rely on the underlying LLM for their performance; however, current literature does not consider the effect of the LLM when comparing these metrics. To account for the effect of LLM, our experiments compare prompt-based metrics across the same set of underlying LLMs.
The major contributions of this work are:

\begin{enumerate}
    \item We introduce AXCEL, a novel prompt-based metric designed to measure the consistency of generated text with respect to a source text. Through extensive experimentation, we demonstrate that AXCEL achieves SOTA performance, outperforming both non-prompt-based and other prompt-based metrics across all tasks.

    \item We show that AXCEL is a generalizable metric for consistency evaluation that is adaptable across multiple tasks and multiple LLMs.    
    \item AXCEL is an explainable metric that provides detailed explanations for the consistency scores and pinpoints the span of text that is inconsistent.
    \item We recalibrated the performance of prompt-based metric baselines by utilizing the latest generation of both open-source and proprietary LLMs. Furthermore, we investigate the role of the underlying LLM in prompt-based metrics.
\end{enumerate}
\section{Related Work}

\paragraph{Non-Prompt Based Metrics:} Non-prompt based metrics primarily consist of textual similarity and NLI based metrics. Textual similarity based metrics rely on a similarity function that computes the similarity between two texts. Methods like ROUGE \citep{lin-2004-rouge} and BLEU \citep{papineni-etal-2002-bleu} use n-gram based lexical graph implementation of the similarity function, whereas, methods like BERTScore \citep{zhang2020bertscore} and MoverScore \citep{zhao2019moverscore} utilize sentence or word embeddings generated by transformers to compute similarity scores. These metrics show poor correlation with human scores \citep{liu-etal-2023-g, zha2023alignscore}. On the other hand, NLI based metrics use entailment models, that compute alignment between a text and a context. These methods either use pre-trained NLI models like Summac \citep{laban2021summac} and custom-trained models like AlignScore \citep{zha2023alignscore}. Metrics based on pre-trained models suffer from poor generalization \citep{mishra-etal-2021-looking}, whereas, ones using custom-trained models have good generalizability but requires a custom training data and consistency scores are not explainable.

\paragraph{Prompt-Based Metrics:} The advent of LLMs has led to an influx of consistency metrics that use LLMs as evaluators. These approaches range from zero-shot evaluators \citep{gao2023humanlike, chen2023evaluating, fu2023gptscore, wang-etal-2023-chatgpt} to those using Chain of Thought (CoT) like G-Eval \citep{liu-etal-2023-g}. These metrics output a score, but they are black-box in nature. The stochastic nature of LLMs and the lack of reasoning behind the scores make these black-box metrics difficult to interpret and use. SelfCheckGPT \cite{manakul2023selfcheckgpt} also includes a prompt-based variant, which performs setence level consitency evaluation and averages sentence level scores to obtain a paragraph level consistency score.
\begin{figure*}[t]
    \centering
    \includegraphics[width=0.8\textwidth]{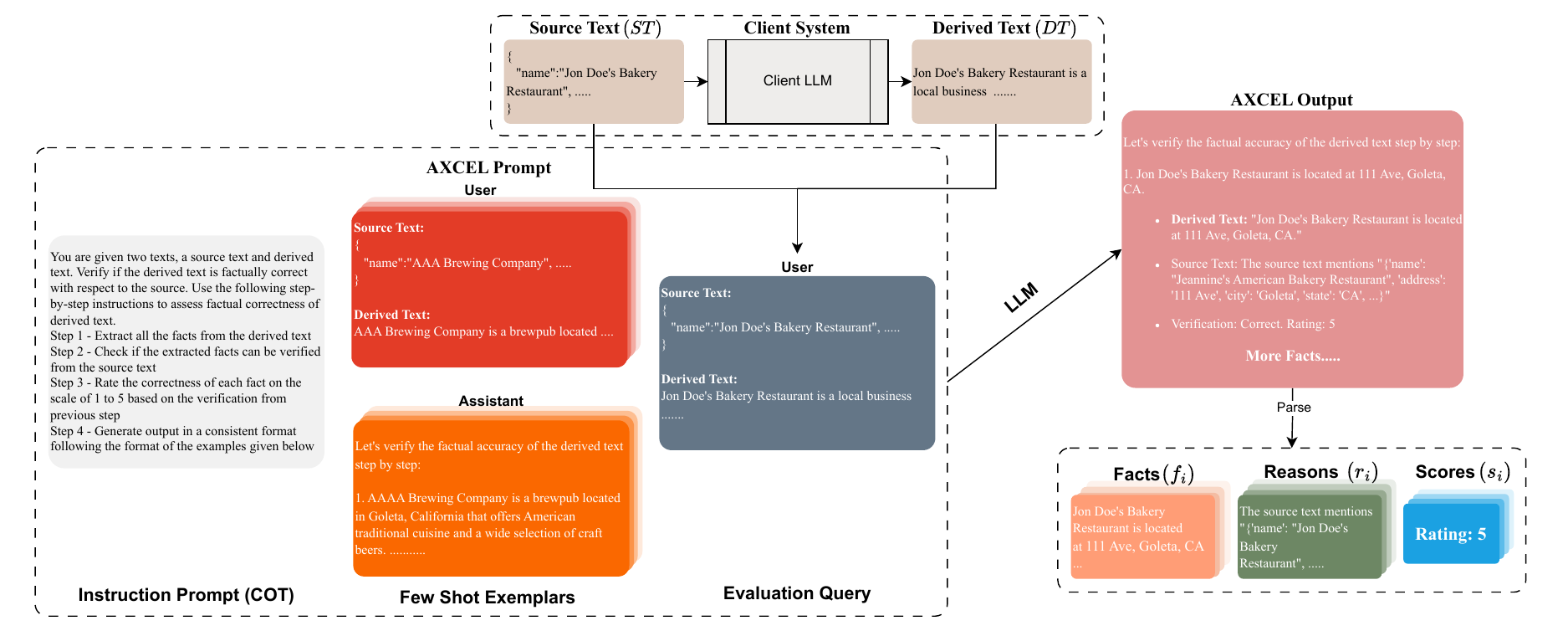}
    \caption{AXCEL computation workflow}
    \label{fig:AXCEL workflow}
\end{figure*}

\section{Methodology}

In this section, consistency is first defined, followed by a detailed discussion of AXCEL's methodology.

\subsection{Consistency}

Consistency of a derived text ($DT$) is computed with respect to a source text ($ST$), where $DT$ is generated by any text generation system like LLM and $ST$ acts as context for $DT$. Consistency is quantified as the ratio of information overlap between $DT$ and $ST$ to the total information in $DT$. This work aims to emulate the process of human evaluation of consistency, which entails verifying the facts present in $DT$ against the information in $ST$. To achieve this, the information in a text $T$, denoted as $I(T)$, is defined as the set of facts, $\{f_i\}$, present in the text. Further, the overlap of a fact $f$ with a text $T$ is defined as $V(f,T)$, where $V$ is a verification function representing the degree to which $f$ is consistent with information in $T$. $V(\cdot,\cdot)$ values range from 1 to 5, where 1 indicates reference to $f$ does not exist in $T$ or $T$ contradicts $f$, and 5 indicates $f$ is completely consistent with $T$. By extension, the consistency of $DT$ with respect to $ST$, $C(DT,ST)$, is defined as:
\begin{equation}
\label{eq:factual_consistency}
    C\left(DT, ST\right) = \frac{\sum_{f \in I\left( DT \right)} V \left( f, ST \right)}{\left|I\left( DT \right) \right|}
\end{equation}
Where $|\cdot|$ represents the set cardinality operation.

\subsection{AXCEL}

\begin{algorithm}
    \caption{AXCEL Methodology Outline}\label{algo:AXCEL}
    \SetKwComment{Comment}{/* }{ */}
    Extract all facts from $DT$, $I\left(DT\right)$\;
    \For{each $f_i$ in $I\left(DT\right)$}{
        \Comment{Verify fact with respect to $ST$, $s_i=V\left(f_i,ST\right)$}
        Locate $S_{DT,i}$ and $S_{ST,i}$, spans of text from $DT$ and $ST$ respectively, where references to fact $f_i$ are present\;
        Assign score $s_i$ between 1 to 5 based on the verification of $S_{DT,i}$ with respect to $S_{ST,i}$ along with the reasoning $r_i$ for the score.     
    }
    Compute overall consistency using equation \ref{eq:factual_consistency}, $C\left(DT,ST\right) \gets \frac{\sum_i s_i}{\left|I\left(DT\right)\right|}$\;
\end{algorithm}

AXCEL computes consistency as defined in Eq. \ref{eq:factual_consistency}. Before delving into AXCEL's prompts, an outline of AXCEL's methodology is provided in Algorithm \ref{algo:AXCEL}. The methodology can be divided into two steps: fact extraction and verification. In the first step, all the facts from the $DT$ are extracted. Subsequently, consistency of these facts are verified using the $ST$, along with reasoning for the consistency scores. Although these are two distinct steps, AXCEL employs a single prompt to achieve both, reducing the number of required LLM calls. The prompt combines, which  chain-of-thought (CoT) and few-shot prompting techniques, can be broken down into three parts: (1) Instruction Prompt, (2) Few-Shot Exemplars, and (3) Evaluation Query. Fig.\ref{fig:AXCEL workflow} illustrates the overall workflow. A detailed exploration of each component is provided in the following subsections.

\subsubsection{Instruction Prompt}

The Instruction Prompt is the CoT component that introduces the task at hand to the LLM. It provides step-by-step instructions to the LLM on how consistency is computed given a pair of $\langle DT, ST 
\rangle$. These instructions are set as a system prompt for the LLM. Complete prompt is provided in Appendix \ref{sec:AXCEL_prompt}

\subsubsection{Few Shot Exemplars}
\label{section: few shot exemplars}

In addition to the instruction prompt, a set of few-shot exemplars are provided to enable in-context learning and improve task understanding. These exemplars also enforce the desired formatting of the LLM response, allowing for easier parsing. The exemplars are structured as triplets, consisting of a $DT$, $ST$, and Response ($R$). $R$ represents the desired output from the LLM given $DT$ and $ST$. It adheres to the steps detailed in Algo. \ref{algo:AXCEL}, i.e., it extracts all the facts and performs verification. The triplets are arranged in a conversational format, with alternating user and assistant blocks: the user part includes the $\langle DT, ST \rangle$ pair, and the assistant part contains $R$.

For each evaluation, that is computation of $C(\cdot,\cdot)$, a set of exemplars, $E_{prompt}$, are randomly sampled from a pool of exemplars, $E_{pool}$, which is a small set of manually annotated triples $\langle DT,ST,R \rangle$. Random sampling ensures that AXCEL's results are not biased by characteristics of a specific $E_{prompt}$. Two types of exemplar creation are possible: in-domain, where $E_{pool}$ is derived from the same domain as the evaluation dataset, and out-of-domain, where $E_{pool}$ is sourced from a domain different from the evaluation dataset. In this work, the primary focus is in-domain exemplars. Using in-domain exemplars introduces a risk of data contamination, where the instance being evaluated might also be selected as part of $E_{prompt}$. To prevent this, it is ensured that the instance being evaluated is not part of $E_{prompt}$ by removing the instance from $E_{pool}$ for that particular evaluation.

\subsubsection{Evaluation Query}
The final part of the prompt communicates the input pair $\langle DT, ST \rangle$ to the LLM for consistency evaluation. The input pair is added as the \textit{user} segment of the conversation. The LLM processes the prompt and generates an output containing $t$ triplets of the form $\langle f_i, r_i, s_i \rangle$, where $f_i$, $r_i$, and $s_i$ represent the $i^{th}$ fact, the reasoning for consistency score, and the consistency score in 1 to 5 range, respectively. The consistency of the input pair is computed as $(\sum_{i}^T s_i)/t$.

\section{Experiments}
In this section, experimental setup and main results are discussed. Due to space constrains additional results, error analysis of AXCEL's explanations, computational cost and detailed prompts are presented in Appendix \ref{sec:appendix}

\subsection{Experimental Setup}
\begin{table}[t]
    \centering    
    \caption{Details of Experimental Design}
    \label{tab:experimental-design}
    \resizebox{\columnwidth}{!}{
        \begin{tabular}{c|c|c|c}
        \toprule
        \textbf{Task}  & \textbf{Datasets} & \textbf{Baselines} & \makecell{\textbf{Evaluation}\\\textbf{Metric}} \\ 
        \midrule
        \makecell{Summar-\\ization} & \makecell{Summeval\\\cite{fabbri2021summeval}, \\ QAGS-CNNDM, \\ QAGS-XSUM\\\cite{wang2020asking}} & \makecell{G-Eval\\\cite{liu-etal-2023-g}, \\ ChatGPT-Eval\\\cite{wang-etal-2023-chatgpt}, \\ AlignScore\\\cite{zha2023alignscore}} & \makecell{Spearman\\Correlation} \\
        \midrule
        \makecell{Free Text\\Generation} & \makecell{WikiBio-GPT-3\\\cite{manakul2023selfcheckgpt}} & \makecell{SelfCheckGPT-Prompt, \\ SelfCheckGPT-NLI\\ \cite{manakul2023selfcheckgpt}} & \makecell{Spearman\\Correlation} \\
        \midrule
        Data2Text & \makecell{RAGTruth\\\cite{wu2023ragtruth}} & \makecell{RAGTruth-Prompt, \\ Finetuned Llama-13B\\\cite{wu2023ragtruth}, \\ AlignScore \\\citep{zha2023alignscore}} & ROC-AUC \\
        \bottomrule
        \end{tabular}
    }
\end{table}

\textbf{Experimental Design:} We assess AXCEL's performance in measuring consistency across three distinct tasks: summarization, free text generation, and converting data to text (Data2Text). For each task, benchmark datasets were identified, along with their corresponding state-of-the-art (SOTA) baselines and evaluation metrics, as detailed in Table \ref{tab:experimental-design}. The selection of these evaluation metrics is based on previous research pertinent to each task. Detailed description of baseline methods and datasets can be found in Appendix \ref{sec:appendix_experimental_setup}

\textbf{Implementation Details:}
In our study, results are reported across three LLMs: two proprietary models (Claude-3-Haiku, and Claude-3-Sonnet), and one open-source model (Llama-3-8B). For consistency in evaluations, other prompt-based metrics were replicated with all the above LLMs using the prompts from their respective papers. For AXCEL, the genreated output was limited to 1000 tokens with a temperature setting of 0. Additional implementation details are provided in Appendix \ref{sec:appendix_implementation_details}. For few-shot prompting, an in-domain exemplar pool ($E_{pool}$) of 10 is randomly sampled from the evaluation dataset. Further, for each evaluation, 3 examples ($E_{prompt}$) were randomly sampled from $E_{pool}$ as the few-shot exemplars.  Ablations on number of exemplars and comparision between in-domain and out-domain exemplars are provided in Section \ref{sec:ablation_studies}

\subsection{Quantitative Results}
\subsubsection{Summarization}

\begin{table*}[t]
    \centering
    \caption{Comparison of Spearman correlation coefficient ($\rho$) between metrics and human evaluation across summary evaluation datasets.}
    \label{tab:summary_results}
    \begin{subtable}{0.95\textwidth}
        \centering
        
        \caption{Comparison of AXCEL against other prompt based metrics across multiple LLMs on summarization evaluation datasets. The last row is the average Spearman correlation of all the summarization evaluation datasets.}
        \label{tab:summary_prompt_results}
        \resizebox{\linewidth}{!}{
        \begin{tabular}{cc|ccc|c}
        \toprule
\textbf{Metric}                       & \textbf{LLM}           & \textbf{SummEval} & \textbf{QAGS-XSUM} & \textbf{QAGS-CNNDM} & \textbf{Average} \\
\midrule
\multirow{5}{*}{\textbf{G-Eval}}      & \textit{Llama-3-8B}    & 39.4              & 33.2               & 38.5                & 37.0             \\
                                      & \textit{Claude-Haiku}  & 44.4              & 21.2               & 51.9                & 39.2             \\
                                      & \textit{Claude-Sonnet} & 59.1              & 56.4               & 67.3                & 60.9             \\
\midrule
\multirow{4}{*}{\textbf{ChatGPTEval}} & \textit{Llama-3-8B}    & 39.4              & 46.1               & 43.6                & 43.0             \\
                                      & \textit{Claude-Haiku}  & 48.1              & 39.0               & 57.1                & 48.0             \\
                                      & \textit{Claude-Sonnet} & 61.2              & 47.9               & 66.9                & 58.7             \\
\midrule
\multirow{4}{*}{\textbf{AXCEL}}       & \textit{Llama-3-8B}    & 47.4              & 56.4               & 67.2                & 57.0             \\
                                      & \textit{Claude-Haiku}  & 57.1              & 58.0               & 68.0                & 61.0             \\
                                      & \textit{Claude-Sonnet} & \textbf{66.4}     & \textbf{62.1}      & \textbf{70.2}       & \textbf{66.2}\\
\bottomrule
\end{tabular}
}
    \end{subtable}
    \begin{subtable}{0.95\textwidth}
        \centering
        
        \caption{Comparison of different types of metrics on summary evaluation datasets. For prompt based metrics, correlations using Claude-Sonnet are displayed in this table. The final column displays the average correlations across these datasets. Textual similarity correlations are quoted from \citep{liu-etal-2023-g}.}
        \label{tab:summary_other_results}
        \resizebox{\linewidth}{!}{
        \begin{tabular}{c|c|ccc|c}
             \toprule
             \textbf{Type} & \textbf{Metric} & \textbf{SummEval} & \textbf{QAGS-XSUM} & \textbf{QAGS-CNNDM} & \textbf{Average} \\
             \midrule
             \multirow{2}{*}{\makecell{Textual\\Similarity}}
             & ROUGE-L & 11.5 & -1.1 & 32.4 & 14.3 \\
             & MoverScore & 15.7 & 4.4 & 34.7 & 18.2 \\
             & BERTScore & 11.0 & 0.8 & 50.5  & 20.8 \\
             & BARTScore & 38.2 & 15.9 & 68 & 40.7 \\
             \midrule
             \multirow{1}{*}{NLI} & AlignScore & 46.6 & 57.2 & \textbf{73.9} & 59.2 \\
             \midrule
             \multirow{3}{*}{\makecell{Prompt\\Based}}
             & G-Eval & 59.1 & 56.4 & 67.3 & 60.9 \\
             & ChatGPTEval & 61.2 & 47.9 & 66.9 & 58.7 \\
             & AXCEL &  \textbf{66.4}     & \textbf{62.1}      & 70.2       & \textbf{66.2} \\
             \bottomrule
        \end{tabular}
        }
    \end{subtable}
\end{table*}

For this task, AXCEL is benchmarked on widely used summary evaluation datasets, SummEval \citep{fabbri2021summeval}, QAGS-CNNDM and QAGS-XSUM \citep{wang2020asking}. The datasets contain pairs of source text and machine generated summary, the goal is to score consistency of the summary with respect to the source.  Consistency of summaries is evaluated using AXCEL by providing the summary as $DT$ and the source as $ST$. First, AXCEL is compared against other prompt-based methods in Table \ref{tab:summary_prompt_results}. All the prompt-based baselines were re-run using the same underlying LLMs as AXCEL to negate the effect of LLM on the performance. In this setting, AXCEL outperforms all the baseline metrics across all LLMs, not only on average but also at the individual dataset level. Unlike G-Eval, which used different prompts for QAGS and SummEval datasets, AXCEL employs the same prompt while improving on performance, showcasing higher generalizability of AXCEL as a metric. This superior performance of AXCEL is attributed to our prompting strategy for two reasons: (1) Our instructions are more detailed and guide the LLM step-by-step in evaluating consistency at a finer granularity, i.e., fact level, whereas other prompt-based metrics instruct the LLM to evaluate at the entire summary level and provide single line description of the task, making the task more difficult and ambiguous; (2) The use of few-shot exemplars with explanations enhances the LLM's understanding of the task and ability to follow instructions via in-context learning. A detailed structural ablation analysis demonstrating the contribution of each component of AXCEL is presented in Section \ref{appendix_structural_analysis}.

Table \ref{tab:summary_other_results} compares Clade-Sonnet variant of prompt-based metrics against other types of metrics. NLI and prompt-based metrics significantly outperform textual similarity-based methods, demonstrating stronger correlation with human judgment. AlignScore outperforms G-Eval and ChatGPTEval across all LLMs, except G-Eval Claude-Sonnet variant. Notably, AXCEL emerges as the SOTA metric, outperforming the best non-prompt-based metric by 11.8\% (66.2 vs 59.2) and the highest-performing baseline prompt-based metric by 8.7\% (66.2 vs 60.9).

\subsubsection{Data2Text}
\begin{table}[t]
    \centering
    
    \caption{Results on RAGTruth Data2Text Benchmark. Here Pr is precision, Re is recall, and AUC is ROC-AUC.}
    \label{tab:data2text results}
    \resizebox{\columnwidth}{!}{
    \begin{threeparttable}[b]
        \begin{tabular}{cc|ccc|c}
             \toprule
             \textbf{Metric} & \textbf{LLM} & \textbf{F1} & \textbf{Pr} & \textbf{Re} & \textbf{AUC} \\
             \midrule
             \multirow{1}{*}{Baseline} & - & 78.3 & 64.3 & \textbf{100} & 50.0 \\
             \midrule
             \multirow{3}{*}{\makecell{RAGTruth \\ Prompt}}
                & Llama-3-8B & 78.4 & 64.5 & 99.8 & 50.2 \\
                & Claude-Haiku & 75.7 & 63.6 & 93.6 & 48.1 \\
                & Claude-Sonnet & 77.9 & 64.4 & 98.6 & 50.1 \\
             \midrule
             \multirow{3}{*}{AXCEL}
                 & Llama-3-8B & 67.8 & 85.7 & 56.1 & 69.7 \\
                 & Claude-Haiku & 70.3 & 84.7 & 60.1 & 70.2 \\
                 & Claude-Sonnet & 85.1 & \textbf{86.2} & 84.1 & \textbf{79.9} \\
             \midrule
             \multirow{1}{*}{Finetune Llama} & Llama-2-13B & \textbf{88.1} & 85.4 & 91 & NA \tnote{1} \\
             \midrule
             \multirow{1}{*}{AlignScore-large} & - & 69.5 & 73.7 & 65.8 & 61.7 \\             
             \bottomrule
        \end{tabular}
    \begin{tablenotes}
        \item[1] {\footnotesize ROC-AUC cannot be computed because the fine-tuned model was not made public. } 
    \end{tablenotes}
    \end{threeparttable}
    }
\end{table}

The dataset for this task, RAGTruth Data2Text \citep{wu2023ragtruth}, comprises LLM generated overviews of JSON data. The objective is to detect hallucinations in these generated overviews. Unlike the other two tasks, which involve only textual data, this task contains both structured (JSON) and textual data, making it a unique application of consistency scoring. AXCEL is compared against prompt-based and fine-tuned Llama-13B metrics described in RAGTruth. The prompt-based metric, provided by RAGTruth, is tailored to this dataset, incorporating details on the types of hallucinations present in the data into the prompt itself. In contract, AXCEL uses the same prompt template and relies on LLM's ability to learn this via in-context learning from the in-domain few shot exemplars.

In the RAGTruth paper, the F1 score is used to assess the effectiveness of metrics in detecting hallucination; however, this metric is affected by skew within the dataset (68\% of generated overviews contain hallucinations). As demonstrated in Table \ref{tab:data2text results}, a simple baseline that classifies all generations as hallucinated achieves a very high F1 score of 78.3, indicating the unsuitability of the F1 score for performance evaluation. To overcome this, we propose using ROC-AUC as the metric, under which this baseline scores 50, indicative of random performance. For AXCEL, JSON is provided as $ST$ and the generated overview as $DT$. Based on AXCEL consistency score, the generated text is marked as hallucinated if consistency score is less than 5 else is marked as not hallucinated. Table \ref{tab:data2text results} compares AXCEL against RAGTruth baselines. AXCEL consistently outperforms the RAGTruth prompt across all model capacities by over 40\% on ROC-AUC. It can also be observed that the RAGTruth prompt is heavily biased towards predicting hallucinations, as shown by scores which are similar to the simple baseline, with disproportionately high recall paired with low precision. Notably, AXCEL with Claude-Sonnet almost matches the performance of the fine-tuned Llama-2-13B model, which was specifically fine-tuned for this dataset, while AXCEL maintains its generality. 

Furthermore, generalizability of AlignScore is tested on this dataset, because the domain of this dataset is different from AlignScore's training data, which primarily contains textual data. We use the publicly available weights of AlignScore-large for this purpose \footnote{\href{https://github.com/yuh-zha/AlignScore/tree/main}{https://github.com/yuh-zha/AlignScore/tree/main}}. The probability estimates from AlignScore were converted to hard labels using a threshold that was obtained through maximizing AUC-ROC on 1000 data points sampled from train split of RAGTruth Data2Text. All variants of AXCEL improve over AlignScore, and the Claude-Sonnet variant outperforms by 29.4\% (79.9 vs 61.7). The margin of improvement achieved by AXCEL over AlignScore has increased in Data2Text task in comparison to Summarization task, showing that AlignScore is not able to generalize to this domain well.

\subsubsection{Free Text Generation}
\begin{table}[t]
    \centering
    \caption{Comparison of $\rho$ between SelfCheckGPT versions and human evaluations for hallucination detection on the WikiBio dataset. Results for non-AXCEL variant are referenced from \citep{manakul2023selfcheckgpt}.}
    \label{tab:hallucination_main_results}
    \resizebox{\columnwidth}{!}{%
        \begin{tabular}{c|cc|c}
            \toprule
             \textbf{Type} & \textbf{Metric} & \textbf{LLM} & $\rho$ \\
             \midrule
             \multirow{1}{*}{NLI} &
             \makecell{SelfCheck\\NLI} & - & 74.1 \\
             \midrule
             \multirow{8}{*}{\makecell{Prompt\\Based}} &
             \multirow{4}{*}{ \makecell{SelfCheck\\Prompt}} & Llama-3-8B & 74.7\\
             & & Claude-Haiku & 77.4\\
             & & Claude-Sonnet & 78.7\\
             \cmidrule{2-4}
             & \multirow{4}{*}{AXCEL} & Llama-3-8B & 80.0 \\
             & & Claude-Haiku & 83.0\\ 
             & & Claude-Sonnet  & \textbf{83.6}\\
             \bottomrule
        \end{tabular}
    }
\end{table}

In this task, hallucinations are detected in free text generations in a zero-resource setting, i.e., without external knowledge. We use WikiBio-GPT3 dataset \cite{manakul2023selfcheckgpt}, which includes passages generated by GPT-3. The approach leverages the framework proposed by SelfCheckGPT that computes a hallucination score in a generation (\(R\)) from an LLM (\(L\)) generated using a prompt (\(P\)) without any external knowledge source. It utilizes the principle of self-consistency, where consistency among multiple generations from the LLM are compared. \(M\) additional generations, (\(G_1, \ldots, G_M\)), are sampled using the same \(L\) and \(P\), and each is compared with \(R\) for consistency. Next, hallucination score, \(H\), is computed by averaging inconsistency scores as follows: 

\begin{equation}
    \label{eq: hallucination}
    H = \frac{1}{M} \sum_{i=1}^M IC\left(R, G_i\right)
\end{equation}
Where $IC(R,G_i)$ is a measure of inconsistency in $R$ with respect to $G_i$. In the context of AXCEL, $IC(R,G_i)$ is computed as $5 - C(R,G_i)$, where 5 represents the upper bound of AXCEL's scoring scale and $R$ is input as $DT$ and $G_i$ as $ST$.

The performance of AXCEL is compared against the prompt-based and NLI-based metrics proposed by SelfCheckGPT. The results in table \ref{tab:hallucination_main_results} demonstrate that AXCEL enhances the performance of SelfCheckGPT across all model capacities, showing improvements of 6.2\% (83.6 vs 78.7) over its prompt-based metric and 12.8\% (83.6 vs 74.1) over its NLI metric. Furthermore, AXCEL reduces the number of LLM calls required to compute consistency scores for a given $R$ and $G_i$ pair compared to SelfCheckGPT. AXCEL requires only one call, whereas SelfCheckGPT's prompt metric, which performs sentence-level evaluation, requires an average of eight calls, as the WikiBio dataset averages about eight sentences per passage.

\subsection{Explainability}
\label{sec:explainability}

\begin{figure}[t]
    \centering
    \includegraphics[width=0.8\columnwidth]{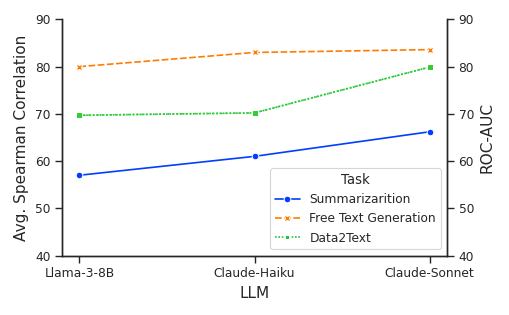}
    \caption{Impact of LLM on prompt based metrics}
    \label{fig:effect_of_llm}
\end{figure}
\begin{figure*}[t]
    \centering
    \includegraphics[width=0.95\textwidth]{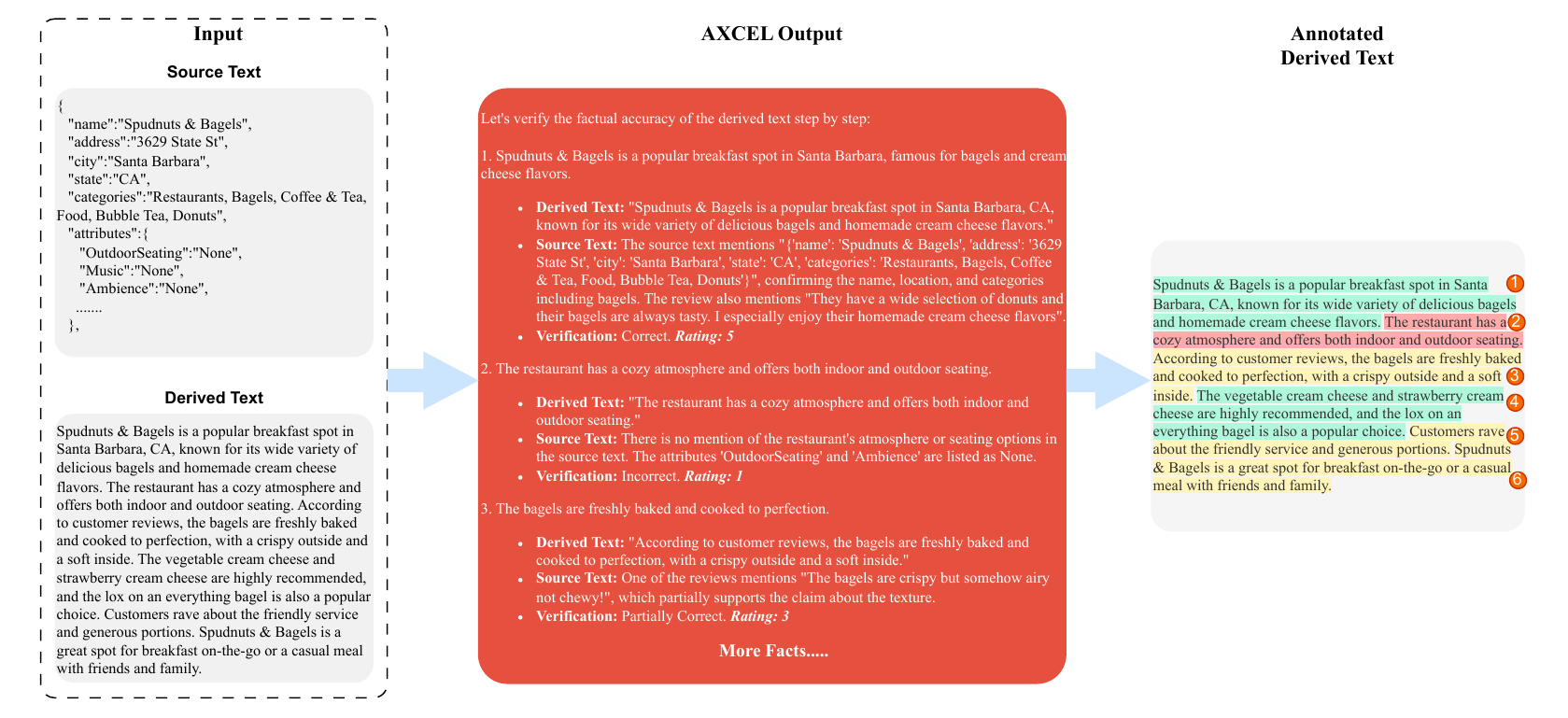}
    \caption{Demonstrating explainability of AXCEL with an example.}
    \label{fig:explainability}
\end{figure*}

Explainability is essential for a metric to build customer trust by clarifying how the score was derived. Current SOTA metrics like G-Eval and AlignScore lack explainability, outputting only a score without any justification, making them black-box methods. This issue is particularly problematic for prompt based metrics because LLMs are prone to hallucination, and relying solely on scores output by these metrics makes it difficult to identify if there are hallucinations in the scoring. AXCEL alleviates this problem and enhances transparency in evaluation by generating scores accompanied by detailed reasoning. Explanations can be used to verify if the generated consistency scores are justifiable, facilitating in easier detection of hallucinated consistency scores.

Fig. \ref{fig:explainability} illustrates an output generated by AXCEL for a $DT$ from the RAGTruth Data2Text dataset. AXCEL extracts six facts from the $DT$, three of which are shown in the illustration for brevity. It scores the first fact as consistent because all the information in the fact can be verified from the $ST$. However, the second fact is inconsistent because the derived text discusses the atmosphere and seating arrangement of the restaurant, but the source JSON contains "None" entries for both outdoor seating and ambience. 
The last fact in the illustration receives a partially consistent score because it could only be partially validated from the source JSON. Additionally, AXCEL's output facilitates the annotation of the derived text, as depicted in the latter part of Fig.\ref{fig:explainability}, with inconsistencies marked in red and consistencies in green. These annotations can serve as valuable feedback for refining the systems responsible for generating such texts.

\subsection{Importance of LLMs in Prompt based Metrics}

LLMs play a crucial role in the effectiveness of prompt-based metrics. To evaluate their impact, we plot the performance of AXCEL as a function of the underlying LLMs across all three tasks, as shown in Figure \ref{fig:effect_of_llm}. 
It can be seen that the performance of AXCEL is a function of the underlying LLM. This pattern holds true for other prompt-based metrics employed in these tasks, with the Claude-Sonnet variant surpassing the performance of all other LLMs, suggesting that improvements in LLMs will likely enhance the performance of these prompt-based metrics in measuring consistency. Therefore, when comparing different prompt-based methods, the effect of the underlying LLM should be taken into consideration.

\subsection{Structural Ablation}
\label{appendix_structural_analysis}
\begin{table}[t]
    \centering
    \caption{Structural ablation: Isolates contribution of each component of AXCEL. AXCEL Claude-Sonnet variant and SummEval datasets are used for this analysis.}
    \label{tab:structural_analysis}
    \resizebox{\columnwidth}{!}{%
        \begin{tabular}{c|c}
            \toprule
             \textbf{Ablation} & \textbf{$\rho$} \\
             \midrule
             Only G-Eval CoT & 59.1\\
             
             Only AXCEL CoT & 64.1\\
             
             Only Few Shot Exemplars (only scores and no explanations) & 53.7\\
             
             Only AXCEL’s Few Shot Exemplars (scores + explanations) & 63.3\\
             
             AXCEL CoT + Few Shot Exemplars (only scores and no explanations) & 63.4\\
             
             \textbf{AXCEL} (AXCEL's CoT + AXCEL's Few Shot Exemplars) & \textbf{66.4}\\

             \bottomrule
        \end{tabular}
    }
\end{table}
To evaluate the individual contributions of AXCEL's components, we conducted a structural ablation study, assessing each component in isolation. Table \ref{tab:structural_analysis} presents the results of this analysis, performed on the Summeval dataset using Claude-Sonnet as the underlying LLM. When employed independently, AXCEL's CoT component demonstrates an 8.4\% improvement over G-Eval's CoT (64.1 vs. 59.1), underscoring the efficacy of our CoT prompt, which provides detailed step-by-step instructions. Similarly, AXCEL's exemplars, which incorporate detailed explanations, outperform those containing only scores by 17.1\% (63.3 vs. 53.7), emphasizing the significance of including explanations within the exemplars which facilitates in-context learning. Notably, both AXCEL's CoT and exemplars individually achieve comparable performance (64.1 and 63.3, respectively). The integration of these components yields the optimal performance, with AXCEL attaining an overall score of 66.4.

\subsection{Exemplar Ablation Studies}
\label{sec:ablation_studies}
To understand the impact of exemplars, we conduct ablation studies using Claude-Haiku as the LLM.

\subsubsection{Effect of number of exemplars}
\label{sec:num_exemplars}

\begin{figure}[t]
    \centering
    \includegraphics[width=\columnwidth]{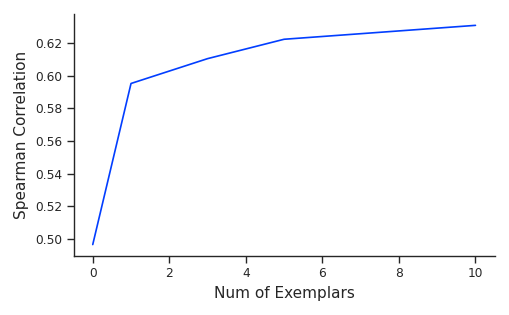} 
    \caption{AXCEL (Claude-Haiku) metric performance varying number of exemplars}
    \label{fig:num_exemplar}
\end{figure}

Figure \ref{fig:num_exemplar} illustrates the impact of number of exemplars used in AXCEL's Claude-Haiku variant on average Spearman correlation across all the summarization datasets. Findings indicate that adding exemplars enhances AXCEL's performance, although the benefits plateau after three exemplars.

\subsubsection{In-Domain vs Out-Domain Exemplars}
\label{sec:indomain_vs_outdomain}
\begin{table}[t]
    \centering
    \caption{Impact of domain of exemplars on AXCEL. Claude-Haiku was used as the underlying LLM for this ablation study.}
    \label{tab:indomain_outdomain}
    \resizebox{\columnwidth}{!}{%
        \begin{tabular}{l|c|c|c|c}
            \toprule
            Dataset & \makecell{Evaluation\\Criteria} & \makecell{AXCEL\\(in-domain)} & \makecell{AXCEL\\(out-domain)} & \makecell{Other\\Prompt Metric} \\
            \midrule
            QAGS-XSUM & Spearman & 58.0 & 56 & \makecell{21.2 (GEval)\\39 (ChatGPT-Eval)} \\
            \midrule
            WikiBio & Spearman & 83 & 82.4 & 78.3 \\
            \midrule
            \makecell{Data2Text\\(RAGTruth)} & ROC-AUC & 70.2 & 57.3 & 48 \\
            \bottomrule
        \end{tabular}
    }
\end{table}

Exemplars play an important role in making AXCEL a generalizable metric across tasks by facilitating in-context learning, helping AXCEL understand the task better. In this ablation study, the impact of using exemplars from domain different to the test setting is examined. Specifically, experiments are conducted for Free Text Generation, QAGS-XSUM, and Data2Text using exemplars from the Summeval datasets, employing Claude-Haiku as the underlying LLM. These datasets were selected because they different from SummEval, making the exemplars markedly different from the test domain. Table \ref{tab:indomain_outdomain} compares AXCEL results using in-domain and out-domain exemplars with other prompt-based metrics. Performance drop is observed when employing out-domain exemplars, which can attributed to increased domain shift between the exemplars and evaluation task. The most significant decline is in data2text, with its JSON inputs causing maximal domain shift. Despite this drop in performance, AXCEL using out-domain exemplars still outperform other prompt-based metrics.
\section{Conclusion}

In this paper, we introduce a novel reference-free, prompt-based metric, AXCEL, designed to measure the consistency of responses generated by LLMs. AXCEL combines the CoT and few-shot prompting techniques to guide LLMs through a process that includes fact extraction and verification. Furthermore, AXCEL offers explainability, providing reasoning for consistency scores, which makes trusting the scores easier as compared to the blackbox approaches. It also helps pin-point the span of hallucinated text which can be used as feedback to improve the upstream text generation system. 

AXCEL was tested on three diverse tasks and was shown to outperform or perform as well as the current SOTA approaches. Non-prompt-based metrics like AlignScore and RAGTruth Llama-2-13B rely on domain-specific training data to achieve generalization, requiring costly data annotation. For instance, AlignScore exhibits suboptimal performance on the RAGTruth Data2Text task. Conversely, prompt-based metrics such as GEval and RAGTruth-prompt employ task-specific prompts, limiting their generalizability. AXCEL bridges these approaches by maintaining consistent instruction prompts across tasks and changing only the few shot exemplars, which requires annotating approximately 10 domain-specific instances, significantly less costly than extensive training or fine-tuning. This highlights AXCEL's ability to generalize as a metric across tasks better than its counterparts.

Furthermore, in this paper, we show that the performance of prompt-based metrics is heavily dependent on the performance of the underlying LLM. With continual improvements in LLMs, the performance of prompt-based metrics is bound to improve. Finally, we show that AXCEL demonstrates strong performance using Llama-3-8B across all three tasks, as seen in tables \ref{tab:summary_prompt_results}, \ref{tab:data2text results} and \ref{tab:hallucination_main_results}; making AXCEL extendable to open source LLMs.

\section*{Limitations}

Given that AXCEL involves an LLM, it is prone to hallucinations. As next steps to improve AXCEL, work needs to be done towards a framework that enables us to quantify the amount of hallucination in each of its fact extraction and verification steps. Further, human annotation of explanations generated by AXCEL needs to collected to access their correctness. AXCEL is more cost-efficient than other prompt-based metrics (check Appendix \ref{sec:computational_cost}), however, reliance of AXCEL on LLMs makes it more computationally intensive than non-prompt based metrics like AlignScore. Another limitation is the usage of in-domain exemplars, which requires manual annotation for each task and domain separately. Our experiments in Appendix \ref{sec:indomain_vs_outdomain} show that AXCEL outperforms other prompt-based metrics using out-domain exemplars, however, performs worse than AXCEL using in-domain variant. Thus, more work has to be done in designing out-domain exemplars that improve the generalization of AXCEL across domains. Finally, the current setup of AXCEL cannot be adopted for tasks where there are two contexts, such as a question-answering setup where the source text and the question are the two contexts. AXCEL needs to be further developed to be able to handle this type of setup.

\bibliography{main}

\begin{thebibliography}{24}
\providecommand{\natexlab}[1]{#1}

\bibitem[{Bang et~al.(2023)Bang, Cahyawijaya, Lee, Dai, Su, Wilie, Lovenia, Ji,
  Yu, Chung, Do, Xu, and Fung}]{bang2023multitask}
Yejin Bang, Samuel Cahyawijaya, Nayeon Lee, Wenliang Dai, Dan Su, Bryan Wilie,
  Holy Lovenia, Ziwei Ji, Tiezheng Yu, Willy Chung, Quyet~V. Do, Yan Xu, and
  Pascale Fung. 2023.
\newblock \href {https://arxiv.org/abs/2302.04023} {A multitask, multilingual,
  multimodal evaluation of chatgpt on reasoning, hallucination, and
  interactivity}.
\newblock \emph{Preprint}, arXiv:2302.04023.

\bibitem[{Brown et~al.(2020)Brown, Mann, Ryder, Subbiah, Kaplan, Dhariwal,
  Neelakantan, Shyam, Sastry, Askell, Agarwal, Herbert-Voss, Krueger, Henighan,
  Child, Ramesh, Ziegler, Wu, Winter, Hesse, Chen, Sigler, Litwin, Gray, Chess,
  Clark, Berner, McCandlish, Radford, Sutskever, and
  Amodei}]{brown2020language}
Tom~B. Brown, Benjamin Mann, Nick Ryder, Melanie Subbiah, Jared Kaplan,
  Prafulla Dhariwal, Arvind Neelakantan, Pranav Shyam, Girish Sastry, Amanda
  Askell, Sandhini Agarwal, Ariel Herbert-Voss, Gretchen Krueger, Tom Henighan,
  Rewon Child, Aditya Ramesh, Daniel~M. Ziegler, Jeffrey Wu, Clemens Winter,
  Christopher Hesse, Mark Chen, Eric Sigler, Mateusz Litwin, Scott Gray,
  Benjamin Chess, Jack Clark, Christopher Berner, Sam McCandlish, Alec Radford,
  Ilya Sutskever, and Dario Amodei. 2020.
\newblock \href {https://arxiv.org/abs/2005.14165} {Language models are
  few-shot learners}.
\newblock \emph{Preprint}, arXiv:2005.14165.

\bibitem[{Chen et~al.(2023)Chen, Gao, and He}]{chen2023evaluating}
Shiqi Chen, Siyang Gao, and Junxian He. 2023.
\newblock Evaluating factual consistency of summaries with large language
  models.
\newblock \emph{arXiv preprint arXiv:2305.14069}.

\bibitem[{Fabbri et~al.(2021)Fabbri, Kry{\'s}ci{\'n}ski, McCann, Xiong, Socher,
  and Radev}]{fabbri2021summeval}
Alexander~R Fabbri, Wojciech Kry{\'s}ci{\'n}ski, Bryan McCann, Caiming Xiong,
  Richard Socher, and Dragomir Radev. 2021.
\newblock Summeval: Re-evaluating summarization evaluation.
\newblock \emph{Transactions of the Association for Computational Linguistics},
  9:391--409.

\bibitem[{Fu et~al.(2023)Fu, Ng, Jiang, and Liu}]{fu2023gptscore}
Jinlan Fu, See-Kiong Ng, Zhengbao Jiang, and Pengfei Liu. 2023.
\newblock \href {https://arxiv.org/abs/2302.04166} {Gptscore: Evaluate as you
  desire}.
\newblock \emph{Preprint}, arXiv:2302.04166.

\bibitem[{Gao et~al.(2023)Gao, Ruan, Sun, Yin, Yang, and
  Wan}]{gao2023humanlike}
Mingqi Gao, Jie Ruan, Renliang Sun, Xunjian Yin, Shiping Yang, and Xiaojun Wan.
  2023.
\newblock \href {https://arxiv.org/abs/2304.02554} {Human-like summarization
  evaluation with chatgpt}.
\newblock \emph{Preprint}, arXiv:2304.02554.

\bibitem[{Hermann et~al.(2015)Hermann, Kociský, Grefenstette, Espeholt, Kay,
  Suleyman, and Blunsom}]{cnndm2015}
Karl~Moritz Hermann, Tomás Kociský, Edward Grefenstette, Lasse Espeholt, Will
  Kay, Mustafa Suleyman, and Phil Blunsom. 2015.
\newblock \href
  {http://papers.nips.cc/paper/5945-teaching-machines-to-read-and-comprehend}
  {Teaching machines to read and comprehend}.
\newblock In \emph{NIPS}, pages 1693--1701.

\bibitem[{Laban et~al.(2021)Laban, Schnabel, Bennett, and
  Hearst}]{laban2021summac}
Philippe Laban, Tobias Schnabel, Paul~N. Bennett, and Marti~A. Hearst. 2021.
\newblock \href {https://arxiv.org/abs/2111.09525} {Summac: Re-visiting
  nli-based models for inconsistency detection in summarization}.
\newblock \emph{Preprint}, arXiv:2111.09525.

\bibitem[{Lin(2004)}]{lin-2004-rouge}
Chin-Yew Lin. 2004.
\newblock \href {https://aclanthology.org/W04-1013} {{ROUGE}: A package for
  automatic evaluation of summaries}.
\newblock In \emph{Text Summarization Branches Out}, pages 74--81, Barcelona,
  Spain. Association for Computational Linguistics.

\bibitem[{Liu et~al.(2023)Liu, Iter, Xu, Wang, Xu, and Zhu}]{liu-etal-2023-g}
Yang Liu, Dan Iter, Yichong Xu, Shuohang Wang, Ruochen Xu, and Chenguang Zhu.
  2023.
\newblock \href {https://doi.org/10.18653/v1/2023.emnlp-main.153} {{G}-eval:
  {NLG} evaluation using gpt-4 with better human alignment}.
\newblock In \emph{Proceedings of the 2023 Conference on Empirical Methods in
  Natural Language Processing}, pages 2511--2522, Singapore. Association for
  Computational Linguistics.

\bibitem[{Manakul et~al.(2023)Manakul, Liusie, and
  Gales}]{manakul2023selfcheckgpt}
Potsawee Manakul, Adian Liusie, and Mark Gales. 2023.
\newblock Selfcheckgpt: Zero-resource black-box hallucination detection for
  generative large language models.
\newblock In \emph{Proceedings of the 2023 Conference on Empirical Methods in
  Natural Language Processing}, pages 9004--9017.

\bibitem[{Mishra et~al.(2021)Mishra, Patel, Vijayakumar, Li, Kapanipathi, and
  Talamadupula}]{mishra-etal-2021-looking}
Anshuman Mishra, Dhruvesh Patel, Aparna Vijayakumar, Xiang~Lorraine Li, Pavan
  Kapanipathi, and Kartik Talamadupula. 2021.
\newblock \href {https://doi.org/10.18653/v1/2021.naacl-main.104} {Looking
  beyond sentence-level natural language inference for question answering and
  text summarization}.
\newblock In \emph{Proceedings of the 2021 Conference of the North American
  Chapter of the Association for Computational Linguistics: Human Language
  Technologies}, pages 1322--1336, Online. Association for Computational
  Linguistics.

\bibitem[{Narayan et~al.(2018)Narayan, Cohen, and Lapata}]{xsum2018}
Shashi Narayan, Shay~B. Cohen, and Mirella Lapata. 2018.
\newblock Don't give me the details, just the summary! topic-aware
  convolutional neural networks for extreme summarization.
\newblock \emph{ArXiv}, abs/1808.08745.

\bibitem[{Ouyang et~al.(2022)Ouyang, Wu, Jiang, Almeida, Wainwright, Mishkin,
  Zhang, Agarwal, Slama, Ray, Schulman, Hilton, Kelton, Miller, Simens, Askell,
  Welinder, Christiano, Leike, and Lowe}]{ouyang2022training}
Long Ouyang, Jeff Wu, Xu~Jiang, Diogo Almeida, Carroll~L. Wainwright, Pamela
  Mishkin, Chong Zhang, Sandhini Agarwal, Katarina Slama, Alex Ray, John
  Schulman, Jacob Hilton, Fraser Kelton, Luke Miller, Maddie Simens, Amanda
  Askell, Peter Welinder, Paul Christiano, Jan Leike, and Ryan Lowe. 2022.
\newblock \href {https://arxiv.org/abs/2203.02155} {Training language models to
  follow instructions with human feedback}.
\newblock \emph{Preprint}, arXiv:2203.02155.

\bibitem[{Papineni et~al.(2002)Papineni, Roukos, Ward, and
  Zhu}]{papineni-etal-2002-bleu}
Kishore Papineni, Salim Roukos, Todd Ward, and Wei-Jing Zhu. 2002.
\newblock \href {https://doi.org/10.3115/1073083.1073135} {{B}leu: a method for
  automatic evaluation of machine translation}.
\newblock In \emph{Proceedings of the 40th Annual Meeting of the Association
  for Computational Linguistics}, pages 311--318, Philadelphia, Pennsylvania,
  USA. Association for Computational Linguistics.

\bibitem[{Rafailov et~al.(2023)Rafailov, Sharma, Mitchell, Ermon, Manning, and
  Finn}]{rafailov2023direct}
Rafael Rafailov, Archit Sharma, Eric Mitchell, Stefano Ermon, Christopher~D.
  Manning, and Chelsea Finn. 2023.
\newblock \href {https://arxiv.org/abs/2305.18290} {Direct preference
  optimization: Your language model is secretly a reward model}.
\newblock \emph{Preprint}, arXiv:2305.18290.

\bibitem[{Raunak et~al.(2021)Raunak, Menezes, and
  Junczys-Dowmunt}]{raunak-etal-2021-curious}
Vikas Raunak, Arul Menezes, and Marcin Junczys-Dowmunt. 2021.
\newblock \href {https://doi.org/10.18653/v1/2021.naacl-main.92} {The curious
  case of hallucinations in neural machine translation}.
\newblock In \emph{Proceedings of the 2021 Conference of the North American
  Chapter of the Association for Computational Linguistics: Human Language
  Technologies}, pages 1172--1183, Online. Association for Computational
  Linguistics.

\bibitem[{Wang et~al.(2020)Wang, Cho, and Lewis}]{wang2020asking}
Alex Wang, Kyunghyun Cho, and Mike Lewis. 2020.
\newblock \href {https://arxiv.org/abs/2004.04228} {Asking and answering
  questions to evaluate the factual consistency of summaries}.
\newblock \emph{Preprint}, arXiv:2004.04228.

\bibitem[{Wang et~al.(2023)Wang, Liang, Meng, Sun, Shi, Li, Xu, Qu, and
  Zhou}]{wang-etal-2023-chatgpt}
Jiaan Wang, Yunlong Liang, Fandong Meng, Zengkui Sun, Haoxiang Shi, Zhixu Li,
  Jinan Xu, Jianfeng Qu, and Jie Zhou. 2023.
\newblock \href {https://doi.org/10.18653/v1/2023.newsum-1.1} {Is {C}hat{GPT} a
  good {NLG} evaluator? a preliminary study}.
\newblock In \emph{Proceedings of the 4th New Frontiers in Summarization
  Workshop}, pages 1--11, Singapore. Association for Computational Linguistics.

\bibitem[{Wu et~al.(2023)Wu, Zhu, Xu, Shum, Niu, Zhong, Song, and
  Zhang}]{wu2023ragtruth}
Yuanhao Wu, Juno Zhu, Siliang Xu, Kashun Shum, Cheng Niu, Randy Zhong, Juntong
  Song, and Tong Zhang. 2023.
\newblock \href {https://arxiv.org/abs/2401.00396} {Ragtruth: A hallucination
  corpus for developing trustworthy retrieval-augmented language models}.
\newblock \emph{Preprint}, arXiv:2401.00396.

\bibitem[{Zha et~al.(2023)Zha, Yang, Li, and Hu}]{zha2023alignscore}
Yuheng Zha, Yichi Yang, Ruichen Li, and Zhiting Hu. 2023.
\newblock \href {https://doi.org/10.18653/v1/2023.acl-long.634}
  {{A}lign{S}core: Evaluating factual consistency with a unified alignment
  function}.
\newblock In \emph{Proceedings of the 61st Annual Meeting of the Association
  for Computational Linguistics (Volume 1: Long Papers)}, pages 11328--11348,
  Toronto, Canada. Association for Computational Linguistics.

\bibitem[{Zhang et~al.(2019)Zhang, Kishore, Wu, Weinberger, and
  Artzi}]{zhang2019bertscore}
Tianyi Zhang, Varsha Kishore, Felix Wu, Kilian~Q Weinberger, and Yoav Artzi.
  2019.
\newblock Bertscore: Evaluating text generation with bert.
\newblock \emph{arXiv preprint arXiv:1904.09675}.

\bibitem[{Zhang et~al.(2020)Zhang, Kishore, Wu, Weinberger, and
  Artzi}]{zhang2020bertscore}
Tianyi Zhang, Varsha Kishore, Felix Wu, Kilian~Q. Weinberger, and Yoav Artzi.
  2020.
\newblock \href {https://arxiv.org/abs/1904.09675} {Bertscore: Evaluating text
  generation with bert}.
\newblock \emph{Preprint}, arXiv:1904.09675.

\bibitem[{Zhao et~al.(2019)Zhao, Peyrard, Liu, Gao, Meyer, and
  Eger}]{zhao2019moverscore}
Wei Zhao, Maxime Peyrard, Fei Liu, Yang Gao, Christian~M. Meyer, and Steffen
  Eger. 2019.
\newblock \href {https://arxiv.org/abs/1909.02622} {Moverscore: Text generation
  evaluating with contextualized embeddings and earth mover distance}.
\newblock \emph{Preprint}, arXiv:1909.02622.

\end{thebibliography}

\appendix

\section{Appendix}
\label{sec:appendix}

\subsection{Kendall-Tau and Pearson Results}
\label{sec:kendall_tau_results}
\begin{table*}[t]
    \centering
    \caption{Comparison of Kendal Tau correlation coefficient ($\tau$) and Pearson correlation ($r$) between metrics and human evaluation across summary evaluation datasets.}
    \label{tab: kendal_tau}    

    \begin{tabular}{c|c|cc|cc|cc|cc}
        \toprule
        \multirow{2}{*}{\textbf{Type}} & \multirow{2}{*}{\textbf{Metric}} & \multicolumn{2}{c}{\textbf{Summeval}}  & \multicolumn{2}{c}{\textbf{QAGS-XSUM}} & \multicolumn{2}{c}{\textbf{QAGS-CNNDM}} & \multicolumn{2}{c}{\textbf{Average}} \\
        \cmidrule{3-10}
        & & $\tau$ & $r$ & $\tau$ & $r$ & $\tau$ & $r$ & $\tau$ & $r$ \\
        \midrule
        \multirow{4}{*}{\makecell{Textual\\Similarity}} & \textbf{ROUGE-L} & 11.5 & - & -0.9 & 2.4 & 25.4 & 35.7 & 12 & - \\
        & MoverScore & 15.7 & - & 3.6 & 5.4 & 27.1 & 41.4 & 15.47 & - \\
        & BERTScore & 11 & - & 0.6 & 2.4 & 39.9 & 57.6 & 17.17 & - \\
        & BARTScore & 31.5 & - & 13 & 18.5 & 55.7 & 73.5 & 33.4 & - \\
        \midrule
        NLI & AlignScore & 37.4 & 66.3 & 46.8 & 52.7 & 61.3 & \textbf{78.1} & 48.5 & 65.7 \\
        \midrule
        \multirow{3}{*}{\makecell{Prompt\\Based}} & Geval & 55.7 & 66.4 & 56.4 & 56.4 & \textbf{62.5} & 65.3 & 58.2 & 62.7 \\
        & ChatGPTEval  & 58.3 & 70.4 & 43.3 & 48.6 & 59.5 & 68.45 & 53.7 & 62.5 \\
        & AXCEL & \textbf{62.6} & \textbf{74.1} & \textbf{57.3} & \textbf{59.4} & 62.1 & 74.9 & \textbf{60.7} & \textbf{69.5} \\
        \bottomrule
    \end{tabular}
\end{table*}
Table \ref{tab: kendal_tau} report the Kendall-Tau correlation and Pearson correlation  numbers for the summarization task. For prompt based metric, we report the numbers using Claude-Sonnet as the LLM, similar to table \ref{tab:summary_other_results}. We observe that AXCEL is able to outperform or perform as well as all the benchmark methods on Kendall-Tau and Pearson as well.

\subsection{Error Analysis}
\begin{table}
    \centering
    \caption{}
    \begin{subtable}{\columnwidth}
        \centering
        \caption{Distribution of AXCEL's errors}
        \label{tab:appendix-error-buckets}
        \begin{tabular}{cc}
        \toprule
        \textbf{Error Bucket} & \textbf{Count}  \\
        \midrule
        Absolute difference is scores \textless{}=1            & 171 \\
        1\textless{}Absolute difference is scores\textless{}=2 & 43   \\
        Absolute difference is scores\textgreater{}2           & 21  \\
        \bottomrule
        \end{tabular}    
    \end{subtable}
    \begin{subtable}{\columnwidth}
        \centering
        \caption{Analysis of Errors}
        \label{tab:error-analysis}
        \begin{tabular}{cc}
            \toprule
            \textbf{Error  Type} & \textbf{Count} \\
            \midrule
            Fact Extraction      & 2              \\
            Verification         & 5              \\
            Scoring              & 6              \\
            Incorrect Label      & 8             \\
            \bottomrule
        \end{tabular}
    \end{subtable}
\end{table}

To understand the accuracy of the explanations generated by AXCEL, an error analysis was performed. The QAGS-CNNDM dataset was categorized into error buckets based on the absolute difference between AXCEL's (using Claude-Sonnet) score and the human-annotated score, as shown in Table \ref{tab:appendix-error-buckets}. The analysis reveals that most instances fall into the $\le 1$ bucket, indicating a high level of agreement between AXCEL's scores and the human annotations.
Further investigation was conducted on the instances with a difference greater than 2 (the $> 2$ bucket), as detailed in Table \ref{tab:error-analysis}. This analysis aimed to identify the steps of AXCEL that are prone to errors. The findings suggest that most of the errors were due to incorrect human annotations, rather than issues with AXCEL's performance.
Interestingly, the number of errors in the different steps of AXCEL seems to be proportional to the complexity of the task in each step. The fact extraction step, being the easiest, had the least number of errors, while the scoring step, the most complex, had the most errors.

\subsection{Experimental Setup Details}
\label{sec:appendix_experimental_setup}

\subsubsection{Dataset Description}
\label{sec:dataset_description}

\paragraph{SummEval \citep{fabbri2021summeval}:} This dataset comprises human annotations from both expert judges and crowdsourced workers, evaluating the consistency and three other aspects of summaries generated by natural text summarization models trained on the CNN/DailyMail news corpus \citep{cnndm2015}. Human evaluators assessed the consistency of summaries generated for 100 documents using 16 different summarization models. To benchmark various consistency metrics, the correlation between the human annotations and the generated metric scores is computed at the document level and averaged across all documents.

\paragraph{QAGS Datasets \citep{wang2020asking}:} QAGS-CNNDM and QAGS-XSUM datasets comprise human annotations evaluating the factual consistency of model-generated summaries for articles from the CNN/DailyMail (CNNDM) \citep{cnndm2015} and XSUM \citep{xsum2018} corpora. QAGS-CNNDM includes multi-line summaries, while QAGS-XSUM contains single-line summaries, referred to as extreme summaries. Human annotators were tasked with marking each sentence in the summary as either consistent or inconsistent with the source article. QAGS-CNNDM and QAGS-XSUM contain 235 and 239 source-summary samples, respectively. The correlation between human annotation scores for summaries and their metric scores is used to evaluate different consistency metrics.

\paragraph{SelfCheckGPT WikiBio \citep{manakul2023selfcheckgpt}:} The dataset serves as a benchmark for evaluating whether generative models exhibit hallucination tendencies. Specifically, the GPT-3 model is utilized to generate Wikipedia articles based on concepts from the WikiBio dataset. Subsequently, human annotations are collected to assess the factual accuracy of the generated passages by comparing them to the ground-truth documents, which are the first paragraphs of actual Wikipedia articles covering the specified concepts. The dataset includes 238 such GPT-3 generations, each accompanied by 20 sampled generated passages. Akin to the QAGS datasets, the correlation between human-annotated hallucination scores and metric-generated hallucination scores is computed to benchmark various methods for evaluating hallucination. 

\paragraph{RAGTruth \citep{wu2023ragtruth}:} The dataset consists of nearly 18,000 generated responses, stemming from three generation tasks: Question Answering, Data-to-text Writing, and News Summarization. In our study, we focus on evaluating the Data-to-text task because it involves JSON data unlike other two tasks. For this task, LLMs are used to generate a text overview of information for randomly selected businesses from the restaurant and nightlife categories of the Yelp Open Dataset. This dataset contains various structured data fields such as BusinessParking, RestaurantsReservations, OutdoorSeating, etc. about restaurants. Additionally, it incorporates up to three business-related user reviews to provide more context. This information is formatted as a JSONs. A total of 1033 JSONs were sampled, with 6 overviews for each JSON using 6 different LLMs (GPT-3.5-turbo-0613, GPT-4-0613, Mistral-7b-Instruct, Llama-2-7B-chat, Llama-2-13B-chat, and Llama-2-70B-chat), culminating in total 6198 responses. Of these responses, 4254 are labelled as containing hallucinations. The dataset further splits this into train and test set, with 900 in test and remaining as train.

\subsubsection{Baseline Description}
\label{sec:baseline_description}
In this section, we describe the baselines used in our study for all the three tasks.
\paragraph{Summarisation:}
\subparagraph{Non-Prompt Based Metrics:}Textual similarity and NLI based metrics are used as representatives of non-prompt based metric comparison. ROUGE-1/2/L \citep{lin-2004-rouge} calculates n-gram overlap between texts to measure the similarity. BERTScore \citep{zhang2019bertscore} determines similarity using token-level embeddings generated by a transformer model. AlignScore \citep{zha2023alignscore} trains a unified entailment model using multiple entailment datasets and utilizes it to compute similarity at the sentence level, which is then aggregated to the passage level. This method currently represents the state-of-the-art among NLI-based metrics.
\subparagraph{Prompt Based Metrics:}GEval \citep{liu-etal-2023-g} and ChatGPT-Eval\citep{wang-etal-2023-chatgpt} are used as baseline. GEval first generates step-by-step evaluation strategies using an Auto Chain of Thought prompt. Subsequently, it prompts LLMs to score the consistency between texts based on the generated steps. ChatGPT-Eval asks the LLM to rate the consitency of the summary with very minimal instructions about the notion of consistency.

\paragraph{Free Text Generation:}
\subparagraph{Non-Prompt Based Metrics:} For comparison we use SelfCheckGPT's \cite{manakul2023selfcheckgpt} NLI, BertScore and Unigram variant. For NLI, it uses a DeBERTa-v3-large fine-tuned to MNLI as the entailment model.
\subparagraph{Prompt Based Metrics:} A zero shot sentence level prompt proposed in the paper is used to measure the consistency.

\paragraph{Data2Text:}
\subparagraph{Non-Prompt Based Metrics:} A Llama-2-13B model was finetuned on the training set of the RAGTruth data, which consists of approximately 15,000 samples. The inputs to the model are the JSON and the response generated by the LLM, and it is trained to output text that contains hallucinations.

\subparagraph{Prompt Based Metrics:} A prompt proposed in RAGTruth\citep{wu2023ragtruth} is used as a baseline. The prompt is optimised for RAGTruth data and contains details on the type of hallucinations that are possible.

\subsection{Implementation Details}
\label{sec:appendix_implementation_details}

For Claude, the model IDs \textit{anthropic.claude-3-haiku-20240307-v1:0} and \textit{anthropic.claude-3-sonnet-20240229-v1:0} were employed for the Claude-Haiku and Claude-Sonnet versions respectively. For Llama-3-8B we used \textit{Llama-3-8B Instruct} in our study.

\subsection{Computational Cost}
\label{sec:computational_cost}
\begin{table*}[t]
\centering
\caption{Token requirement of each metric across all the summarization datasets}
\label{tab:token_requirement}
\begin{tabular}{c|cc|cc|cc}
\toprule
\multirow{2}{*}{\textbf{Dataset}} & \multicolumn{2}{c}{\textit{\textbf{AXCEL}}}                                    & \multicolumn{2}{c}{\textit{\textbf{GEval}}}                                    & \multicolumn{2}{c}{\textit{\textbf{ChatGPTEval}}}                              \\
\cmidrule{2-7}
                         & \makecell{Avg. Input\\Tokens} & \makecell{Avg. Output\\Tokens} & \makecell{Avg. Input\\Tokens} & \makecell{Avg. Output\\Tokens} & \makecell{Avg. Input\\Tokens} & \makecell{Avg. Output\\Tokens} \\
\midrule
\textbf{Summeval}                 & 3700.23                               & 365.63                                 & 799                                   & 5                                      & 672                                   & 5                                      \\
\textbf{XSUM}                     & 2582.88                               & 187.63                                 & 554                                   & 5                                      & 579                                   & 5                                      \\
\textbf{CNNDM}                    & 3172.94                               & 368.34                                 & 542                                   & 5                                      & 567                                   & 5                                     \\
\bottomrule
\end{tabular}
\end{table*}

\begin{table*}[t]
\centering
\caption{Computation cost comparision of AXCEL (Claude-Haiku) and prompt-based metric baseline using Claude-Sonnet which have comparable performance. Each row has been scaled with respect to AXCEL (Claude-Haiku) cost, that is, each row shows the cost relative to AXCEL's Claude-Haiku variant cost.}
\label{tab:cost_comparison}
\begin{tabular}{c|ccc}
\toprule
\textbf{Dataset} & \textbf{AXCEL (Claude-Haiku)} & \textbf{GEval (Claude-Sonnet)} & \textbf{ChatGPTEval (Claude-Sonnet)} \\
\midrule
\textbf{Summeval}         & 1                                & 1.79                                  & 1.51                                      \\
\textbf{XSUM}             & 1                               & 1.98                                  & 2.06                                        \\
\textbf{CNNDM}            & 1                               & 1.36                                  & 1.42  \\
\bottomrule
\end{tabular}
\end{table*}

Table \ref{tab:token_requirement} provides an analysis of the token requirements for each prompt-based metric in summarization tasks. It can be observed that AXCEL requires a higher number of input and output tokens compared to GEval and ChatGPTEval. This increase is due to the use of few-shot exemplars in AXCEL, which contribute to the high input token count. Additionally, AXCEL's output includes not only a score but also detailed reasoning, resulting in a higher output token count—a tradeoff necessary for enhanced explainability. However, this same token count allows AXCEL to perform effectively with smaller models. Specifically, AXCEL, when used with the Claude-Haiku/Llama-3-8B model, outperforms/matches both GEval and ChatGPTEval using the Claude-Sonnet model, as illustrated in Table \ref{tab:summary_prompt_results}. This allows AXCEL to be a cost-efficient metric, as shown in Table \ref{tab:cost_comparison}.

\subsection{AXCEL Prompt}
\label{sec:AXCEL_prompt}
The following sections outline the prompts utilized for all LLMs. The only different between prompts used for Claude family LLMs and other LLMs is the use of $XML$ tags for Claude models. The prompts presented use only a single exemplar for its few-shot prompting.

\subsubsection{LLama prompt}
\label{sec:gpt_prompt}

\textbf{System: } \texttt{You are given two texts, a source text and derived text. Verify if the derived text is factually correct with respect to the source. Use the following step-by-step instructions to assess factual correctness of derived text. \\ Step 1 - Extract all the facts from the derived text. \\ Step 2 - Check if the extracted facts can be verified from the source text. \\ Step 3 - Rate the correctness of each fact on the scale of 1 to 5 based on the verification from previous step. \\ Step 4 - Generate output in a consistent format.}

\textbf{User: } \texttt{Source Text:
Manchester City are keen to sign Anderlecht teenager Evangelos Patoulidis. The 14-year-old playmaker is regarded as one of the best talents to emerge from Anderlecht's youth set-up and has also attracted attention from Arsenal and Barcelona. The Belgian starlet rejected a move to Barcelona's La Masia academy when he was 12 as his family wanted him to continue his studies . He has continued to impress and City have held discussions with Anderlecht chairman Roger Vanden Stock in the hope of agreeing a compensation package. Manuel Pellegrini is looked to build for the future by snapping up hot property Evangelos Patoulidis.}

\texttt{Derived Text: \\
Evangelos patoulidis is regarded as one of the best players to emerge from anderlecht youth. He has also attracted attention from arsenal and barcelona. The belgian starlet rejected a move to barcelona 's la masia academy. The 14-year-old has attracted interest from barcelona to barcelona.}

\textbf{Assistant: } \texttt{Let's verify the factual accuracy of the derived text step by step:}

\texttt{1. Evangelos Patoulidis is Regarded as One of the Best Players to Emerge from Anderlecht Youth:\\- **Derived Text:** Evangelos Patoulidis is regarded as one of the best players to emerge from Anderlecht youth.\\- **Source Text:** The source text states that Patoulidis is regarded as "one of the best talents to emerge from Anderlecht's youth set-up".\\- **Verification:** Correct. Rating: 5}

\texttt{2. He Has Also Attracted Attention from Arsenal and Barcelona:\\- **Derived Text:** He has also attracted attention from Arsenal and Barcelona.\\- **Source Text:** This fact is mentioned verbatim in the source text.\\- **Verification:** Correct. Rating: 5}

\texttt{3. The Belgian Starlet Rejected a Move to Barcelona's La Masia Academy:\\- **Derived Text:** The Belgian starlet rejected a move to Barcelona's La Masia academy.\\- **Source Text:** The source text confirms this fact.\\- **Verification:** Correct. Rating: 5}

\texttt{4. The 14-Year-Old Has Attracted Interest from Barcelona to Barcelona:\\- **Derived Text:** The 14-year-old has attracted interest from Barcelona to Barcelona.\\- **Source Text:** This statement is confusing and not supported by the source text.\\- **Verification:** Incorrect. Rating: 1}

\textbf{User: } \texttt{Source Text: \\
\{source\_text\} \\
\\
Derived Text: \\
\{derived\_text\}}

\textbf{Assistant: }

\subsubsection{Claude prompt}
\label{sec:claude_prompt}
\textbf{System: } \texttt{You are given two texts, a source text and derived text. Verify if the derived text is factually correct with respect to the source. Use the following step-by-step instructions to assess factual correctness of derived text. \\ Step 1 - Extract all the facts from the derived text. \\ Step 2 - Check if the extracted facts can be verified from the source text. \\ Step 3 - Rate the correctness of each fact on the scale of 1 to 5 based on the verification from previous step. \\ Step 4 - Generate output in a consistent format following the format of the examples given below.}

\textbf{User: } \texttt{<Source Text>:
Manchester City are keen to sign Anderlecht teenager Evangelos Patoulidis. The 14-year-old playmaker is regarded as one of the best talents to emerge from Anderlecht's youth set-up and has also attracted attention from Arsenal and Barcelona. The Belgian starlet rejected a move to Barcelona's La Masia academy when he was 12 as his family wanted him to continue his studies . He has continued to impress and City have held discussions with Anderlecht chairman Roger Vanden Stock in the hope of agreeing a compensation package. Manuel Pellegrini is looked to build for the future by snapping up hot property Evangelos Patoulidis. \\
</Source Text>}

\texttt{<Derived Text> \\
Evangelos patoulidis is regarded as one of the best players to emerge from anderlecht youth. He has also attracted attention from arsenal and barcelona. The belgian starlet rejected a move to barcelona 's la masia academy. The 14-year-old has attracted interest from barcelona to barcelona. \\
</Derived Text>
}

\textbf{Assistant: } \texttt{Let's verify the factual accuracy of the derived text step by step:}

\texttt{1. Evangelos Patoulidis is Regarded as One of the Best Players to Emerge from Anderlecht Youth:\\- **Derived Text:** Evangelos Patoulidis is regarded as one of the best players to emerge from Anderlecht youth.\\- **Source Text:** The source text states that Patoulidis is regarded as "one of the best talents to emerge from Anderlecht's youth set-up".\\- **Verification:** Correct. Rating: 5}

\texttt{2. He Has Also Attracted Attention from Arsenal and Barcelona:\\- **Derived Text:** He has also attracted attention from Arsenal and Barcelona.\\- **Source Text:** This fact is mentioned verbatim in the source text.\\- **Verification:** Correct. Rating: 5}

\texttt{3. The Belgian Starlet Rejected a Move to Barcelona's La Masia Academy:\\- **Derived Text:** The Belgian starlet rejected a move to Barcelona's La Masia academy.\\- **Source Text:** The source text confirms this fact.\\- **Verification:** Correct. Rating: 5}

\texttt{4. The 14-Year-Old Has Attracted Interest from Barcelona to Barcelona:\\- **Derived Text:** The 14-year-old has attracted interest from Barcelona to Barcelona.\\- **Source Text:** This statement is confusing and not supported by the source text.\\- **Verification:** Incorrect. Rating: 1}

\textbf{User: } \texttt{<Source Text> \\
\{source\_text\} \\
</Source Text>} \\

\texttt{<Derived Text> \\
\{derived\_text\} \\
</Derived Text>}

\textbf{Assistant: }

\end{document}